# Upgrading Ambiguous Signs in QPNs

Janneke H. Bolt, Silja Renooij, and Linda C. van der Gaag
Institute of Information and Computing Sciences, Utrecht University
P.O. Box 80.089, 3508 TB Utrecht, The Netherlands
{janneke,silja,linda}@cs.uu.nl

**Abstract**

A qualitative probabilistic network models the probabilistic relationships between its variables by means of signs. Non-monotonic influences have associated an ambiguous sign. These ambiguous signs typically give rise to uninformative results upon inference. We argue that a non-monotonic influence can be associated with a more informative sign that indicates its effect in the current state of the network. To capture this effect, we introduce the concept of situational sign. Furthermore, if the network converts to a state in which all variables that provoke the non-monotonicity have been observed, a non-monotonic influence reduces to a monotonic one. We study the persistence and propagation of situational signs upon inference and give a method for establishing the sign of a reduced influence.

## 1 INTRODUCTION

The formalism of Bayesian networks is generally considered an intuitively appealing and powerful formalism for capturing complex problem domains along with their uncertainties. The usually large number of probabilities required for a Bayesian network, however, tends to pose a major obstacle to their construction [1]. Research on facilitating probability assessment for Bayesian networks has benefited to some extent from the concept of *qualitative probabilistic network* (QPN) [2]. Like a Bayesian network, a qualitative network encodes statistical variables and the probabilistic relationships between them in a directed acyclic graph. The relationships between the variables are not quantified by conditional probabilities as in a Bayesian network but are summarised by qualitative signs instead [3]. For inference with such a qualitative network, an efficient algorithm is available, based on the idea of propagating and combining signs [4].

Although qualitative probabilistic networks do not require numerical probabilities, their high level of abstraction brings some disadvantages of its own. Among these is the fact that qualitative networks do not provide for an informative way of capturing probabilistic influences that are positive in one state and negative in another state of the network. Such non-monotonic influences are associated with the *ambiguous sign* '?'. These ambiguous signs typically lead to ambiguous, and therefore uninformative, results upon inference. In this paper we propose to upgrade the ambiguous signs associated with non-monotonic influences with additional information and we show how this information can be exploited to forestall, to some extent, ambiguous results upon inference.

We extend the framework of qualitative probabilistic networks with the concept of *situational sign*. This concept is motivated by the observation that in each state of a network, any non-monotonic influence is unambiguous. A situational sign now is associated with a non-monotonic influence and captures information about the effect of the influence in the current state of the network. Since a situational sign depends on the network's state, we investigate its persistence to changes in the current state and provide a method for its updating if necessary. We further argue that a network may convert to a state in which all variables that underlie the non-monotonicity of an influence, its so-called *provokers*, have been observed. In such a state, the non-monotonic influence under study reduces to a fixed monotonic one. We adapt the standard propagation algorithm to provide for inference with a qualitative network with situational signs and extend it by a method for establishing signs for reduced influences.

The remainder of the paper is organised as follows. Section 2 provides some preliminaries on qualitative probabilistic networks. Section 3 introduces our concepts of situational sign and provoker. Section 4 presents a sign-propagation algorithm that is adapted to qualitative networks with situational signs and extended to exploit the observation of provokers. The paper is rounded off with some conclusions and directions for further research in Section 5.

## 2 PRELIMINARIES

A Bayesian network is a concise representation of a joint probability distribution Pr on a set of statistical variables. In the sequel, (sets of) variables are denoted by upper-case letters. For ease of exposition, we assume all variables to be binary, writing $a$ for $A = \textit{true}$ and $\bar{a}$ for $A = \textit{false}$; we further assume that $a > \bar{a}$. Each variable is represented by a node in a directed acyclic graph $G$. The probabilistic relationships between the variables are captured by the arcs $A(G)$ of the digraph. Associated with each variable $A$ is a set of conditional probability distributions $\Pr(A \mid \pi(A))$, with $\pi(A)$ the set of parents of $A$ in $G$. We introduce a small Bayesian network for our running example.

**Example 1** The Bayesian network from Figure 1 represents a fragment of fictitious knowledge about the effect of training and fitness on a feeling of well-being. Node $T$ models whether or not one has undergone a training session, node $F$ captures one's fitness, and node $W$ models whether or not one is feeling well. □

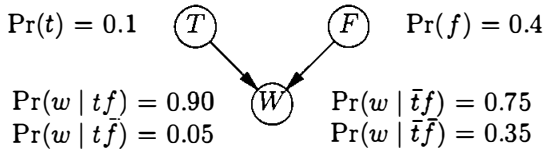

$\Pr(t) = 0.1$    $T$    $F$    $\Pr(f) = 0.4$

$\Pr(w \mid tf) = 0.90$    $W$    $\Pr(w \mid \bar{t}f) = 0.75$
$\Pr(w \mid t\bar{f}) = 0.05$        $\Pr(w \mid \bar{t}\bar{f}) = 0.35$

Figure 1: An example Bayesian network, modelling the influences of training ($T$) and fitness ($F$) on a feeling of well-being ($W$).

In its initial state, where no evidence has been entered, a Bayesian network captures a prior probability distribution over its variables. As observations become available, the network converts to another state, representing the posterior distribution given the evidence.

*Qualitative probabilistic networks* were introduced as qualitative abstractions of Bayesian networks: instead of conditional probability distributions, a qualitative probabilistic network associates with its digraph qualitative influences and qualitative synergies [3]. A *qualitative influence* between two nodes expresses how the values of one node influence the probabilities of the values of the other node. For example, a *positive qualitative influence* of a node $A$ on a node $B$, denoted $S^+(A, B)$, expresses that observing a high value for $A$ makes the higher value for $B$ more likely, regardless of any other direct influences on $B$, that is

$$\Pr(b \mid ax) - \Pr(b \mid \bar{a}x) \geq 0$$

for any combination of values $x$ for the set $\pi(B) \setminus \{A\}$ of parents of $B$ other than $A$. A negative qualitative influence, denoted $S^-$, and a zero qualitative influence, denoted $S^0$, are defined analogously. A non-monotonic or unknown influence of node $A$ on node $B$ is denoted by $S^?(A, B)$.

The set of all influences of a qualitative network exhibits various important properties [3]. The property of *symmetry* states that, if the network includes the influence $S^\delta(A, B)$, then it also includes $S^\delta(B, A)$, $\delta \in \{+, -, 0, ?\}$. The *transitivity* property asserts that the signs of qualitative influences along a trail without head-to-head nodes combine into a sign for the net influence with the $\otimes$-operator from Figure 2. The property of *composition* asserts that the signs of multiple influences between nodes along parallel trails combine into a sign for the net influence with the $\oplus$-operator.

| $\otimes$ | + | − | 0 | ? | | $\oplus$ | + | − | 0 | ? |
|---|---|---|---|---|---|---|---|---|---|---|
| + | + | − | 0 | ? | | + | + | ? | + | ? |
| − | − | + | 0 | ? | | − | ? | − | − | ? |
| 0 | 0 | 0 | 0 | 0 | | 0 | + | − | 0 | ? |
| ? | ? | ? | 0 | ? | | ? | ? | ? | ? | ? |

Figure 2: The $\otimes$- and $\oplus$-operators.

A qualitative probabilistic network further includes *additive synergies*. An additive synergy expresses how two nodes interact in their influence on a third node. For example, a *positive additive synergy* of a node $A$ and a node $B$ on a common child $C$, denoted $Y^+(\{A, B\}, C)$, expresses that the joint influence of $A$ and $B$ on $C$ exceeds the sum of their separate influences regardless of any other direct influences on $C$, that is,

$$\Pr(c \mid abx) + \Pr(c \mid \bar{a}\bar{b}x) \geq \Pr(c \mid a\bar{b}x) + \Pr(c \mid \bar{a}bx)$$

for any combination of values $x$ for the set $\pi(C) \setminus \{A, B\}$ of parents of $C$ other than $A$ and $B$. A negative additive synergy, denoted $Y^-$, and a zero additive synergy, denoted $Y^0$, are defined analogously. A non-monotonic or unknown additive synergy of $A$ and $B$ on $C$ is denoted by $Y^?(\{A, B\}, C)$.

**Example 2** We consider the qualitative abstraction of the Bayesian network from Figure 1. From the conditional probability distributions specified for node $W$, we have that $\Pr(w \mid tf) - \Pr(w \mid t\bar{f}) \geq 0$ and $\Pr(w \mid \bar{t}f) - \Pr(w \mid \bar{t}\bar{f}) \geq 0$, and therefore that $S^+(F, W)$: fitness favours well-being regardless of training. We further have that $\Pr(w \mid tf) - \Pr(w \mid \bar{t}f) > 0$ and $\Pr(w \mid t\bar{f}) - \Pr(w \mid \bar{t}\bar{f}) < 0$, and therefore that $S^?(T, W)$: the effect of training on well-being depends on one's fitness. From $\Pr(w \mid tf) + \Pr(w \mid \bar{t}\bar{f}) \geq \Pr(w \mid t\bar{f}) + \Pr(w \mid \bar{t}f)$, to conclude, we find that $Y^+(\{T, F\}, W)$. The resulting qualitative network is shown in Figure 3; the signs of the influences are shown along the digraph's arcs, and the sign of the additive synergy is indicated over the curve over node $W$. □

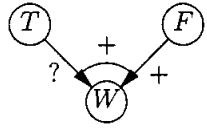

Figure 3: The qualitative abstraction of the Bayesian network from Figure 1.

We would like to note that, although in the previous example the qualitative relationships between the variables are computed from the conditional probabilities of the corresponding quantitative network, in real applications these relationships are elicited directly from domain experts.

For inference with a qualitative probabilistic network, a polynomial-time algorithm based on the idea of propagating and combining signs is available [4]. This algorithm traces the effect of observing a value for a node on the other nodes in a network by message-passing between neighbouring nodes. The algorithm is summarised in pseudo-code in Figure 4. For each node $V$, a *node sign* 'sign$[V]$' is determined, indicating the direction of change in its probability distribution that is occasioned by the new observation; initial node signs equal '0'. Observations are entered as a '+' for the observed value *true*, or a '−' for the value *false*. Each node receiving a message updates its sign using the ⊕-operator and subsequently sends a message to each neighbour that is not independent of the observed node. The sign of this message is the ⊗-product of the node's (new) sign and the sign of the influence it traverses. This process is repeated throughout the network, building on the properties of symmetry, transitivity, and composition of influences.

## 3 UPGRADING '?'s

The presence of non-monotonic influences with the ambiguous sign '?' in a qualitative probabilistic network is likely to give rise to ambiguous, and therefore uninformative, results upon inference. In this section we study the non-monotonicity of influences and argue that ambiguous results upon inference can be forestalled to some extent. In Section 3.1 we introduce the concept of *situational sign* to capture the effect of a non-monotonic influence in the current state of the network. In Section 3.2 we associate with a non-monotonic influence the set of variables that, upon ob-

**procedure** Process-Obs($Q, Obs, sign$):
  **for all** $V_i \in V(G)$ in $Q$
  **do** sign$[V_i] \leftarrow$ '0';
  Propagate-Sign($Q, \varnothing, Obs, sign$).

**procedure** Propagate-Sign($Q, trail, to, message$):
  sign$[to] \leftarrow$ sign$[to] \oplus message$;
  $trail \leftarrow trail \cup \{to\}$;
  **for** each neighbour $V_i$ of $to$ in $Q$
  **do** $linksign \leftarrow$ sign of influence between $to$ and $V_i$;
    $message \leftarrow$ sign$[to] \otimes linksign$;
    **if** $V_i \notin trail$ and sign$[V_i] \neq$ sign$[V_i] \oplus message$
    **then** Propagate-Sign($Q, trail, V_i, message$).

Figure 4: The sign-propagation algorithm.

servation, serve to reduce the non-monotonic influence to a fixed monotonic one.

### 3.1 SITUATIONAL SIGNS

In this section we introduce the concept of *situational sign* into the formalism of qualitative probabilistic networks to capture additional information about the current effect of a non-monotonic influence. We show that, upon inference, such situational signs can be propagated and combined as regular qualitative signs. As situational signs are dynamic in nature, we give a method for maintaining their validity as evidence is being entered.

#### 3.1.1 Definition

We recall that a qualitative influence of a node $A$ on a node $B$ is monotonic if the difference $\Pr(b \mid ax) - \Pr(b \mid \bar{a}x)$ has the same sign for *all* combinations of values $x$ for the set $X = \pi(B) \setminus \{A\}$. This sign then constitutes the unambiguous sign of the influence. We note that this sign is not just valid for all combinations of values $x$, but also given any distribution $\Pr(X)$ over these combinations of values. If the difference $\Pr(b \mid ax) - \Pr(b \mid \bar{a}x)$ yields contradictory signs for different combinations of values $x$, however, the influence of $A$ on $B$ is non-monotonic and is assigned the ambiguous sign '?'. Yet, in each specific state of the network, associated with a (possibly posterior) probability distribution $\Pr(X)$, the influence of $A$ on $B$ is unambiguous, that is, either positive, negative or zero. To capture the sign of a non-monotonic influence in a specific state of the network, we now introduce the concept of situational sign. We use the notation $[\Pr(b \mid a) - \Pr(b \mid \bar{a})]_{\Pr(X)}$ to denote the difference between $\Pr(b \mid a)$ and $\Pr(b \mid \bar{a})$ in the state of the network associated with $\Pr(X)$.

**Definition 1** *Let $G$ be the digraph of a qualitative probabilistic network, with $A, B \in V(G)$ and $A \to B \in$*

$A(G)$. Let $X = \pi(B) \setminus \{A\}$ and let $\Pr(X)$ be a joint probability distribution over $X$. An influence of node $A$ on node $B$ with a positive situational sign given $\Pr(X)$, denoted $S^{?(+)x}(A,B)$, expresses that

- $S^?(A,B)$, and
- $[\Pr(b \mid a) - \Pr(b \mid \bar{a})]_{\Pr(X)} \geq 0$

A negative situational sign given $\Pr(X)$, denoted '$?(-)_X$', and a zero situational sign given $\Pr(X)$, denoted '$?(0)_X$', have analogous meanings. An unknown situational sign given $\Pr(X)$ is denoted '$?(?)_X$'. In the sequel, an influence with a situational sign will be called a *situational influence*. A qualitative network extended with situational signs will be termed a *situational qualitative network*. Note that while the signs of regular qualitative influences have general validity, situational signs are dynamic in nature and pertain to a specific state of the network.

**Example 3** Consider once again the example Bayesian network from Figure 1 and its qualitative abstraction shown in Figure 3. The qualitative influence of node $T$ on node $W$ was found to be non-monotonic. Its sign therefore depends on the state of the network. In the prior state of the network, where no evidence has been entered, we have that $\Pr(f) = 0.4$. Given the prior probability distribution over node $F$, we compute $\Pr(w \mid t) = 0.39$ and $\Pr(w \mid \bar{t}) = 0.51$. From the difference $\Pr(w \mid t) - \Pr(w \mid \bar{t}) = -0.12$ being negative, we conclude that the current influence of node $T$ on node $W$ is negative. The situational sign of the influence is therefore '$?(-)_{\{F\}}$'. The situational qualitative network for the prior state is shown in Figure 5. □

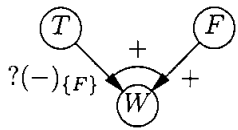

Figure 5: The network from Figure 3, now with the prior situational influence of $T$ on $W$.

Once again we note that, although in the previous example the sign of the situational influence is computed from the quantitative network, in a real application it would be elicited directly from a domain expert. In the remainder of the paper, we assume that the expert has given situational signs for all non-monotonic influences for the prior state of the network.

### 3.1.2 Properties

The sign-propagation algorithm for inference with a qualitative probabilistic network builds upon the properties of symmetry, transitivity and composition of qualitative influences. In this section we argue that situational influences exhibit these properties as well, thereby providing for the propagation of situational signs upon inference.

**Proposition 1** Let $G$ be the digraph of a situational network, with $A, B \in V(G)$ and $A \to B \in A(G)$. Let $X = \pi(B) \setminus \{A\}$. For all $\delta \in \{+, -, 0, ?\}$, if $S^{?(\delta)x}(A,B)$, then $S^{?(\delta)x}(B,A)$.

**Proof.** We outline the proof for $\delta = +$; similar arguments apply to the other signs. We have that

$$S^{?(+)x}(A,B) \Rightarrow [\Pr(b \mid a) - \Pr(b \mid \bar{a})]_{\Pr(X)} \geq 0 \wedge$$
$$\exists x \, \Pr(b \mid ax) - \Pr(b \mid \bar{a}x) > 0 \wedge$$
$$\exists x \, \Pr(b \mid ax) - \Pr(b \mid \bar{a}x) < 0$$

By multiplying the left and right hand sides of the first inequality above with $\frac{\Pr(a)\cdot\Pr(\bar{a})}{\Pr(b)\cdot\Pr(\bar{b})}$, and those of the second and third inequality above with $\frac{\Pr(ax)\cdot\Pr(\bar{a}x)}{\Pr(bx)\cdot\Pr(\bar{b}x)}$, analogous to the proof of symmetry of a regular qualitative influence [5], we find that

$$S^{?(+)x}(A,B) \Rightarrow [\Pr(a \mid b) - \Pr(a \mid \bar{b})]_{\Pr(X)} \geq 0 \wedge$$
$$\exists x \, \Pr(a \mid bx) - \Pr(a \mid \bar{b}x) > 0 \wedge$$
$$\exists x \, \Pr(a \mid bx) - \Pr(a \mid \bar{b}x) < 0$$

We conclude that

$$S^{?(+)x}(A,B) \Rightarrow S^{?(+)x}(B,A)$$

□

The property of transitivity of regular qualitative influences guarantees that for computing the effect of a change in a node's probability distribution on the marginal distributions over its neighbours, it is valid to propagate the change over the incident influences using the $\otimes$-operator from Figure 2. The sign-propagation algorithm for inference with a regular qualitative network in fact builds on this property [4]. We now show that the same property holds for situational influences. In doing so, we will use the notation $\text{sign}[B]^{A \to B}$ to denote the direction of change in the current probability distribution over $B$ that is occasioned by a change in the distribution over $A$ along the arc $A \to B$.

**Proposition 2** Let $G$ be the digraph of a situational network, with $A, B \in V(G)$ and $A \to B \in A(G)$. Let $X = \pi(B) \setminus \{A\}$ and let $S^{?(\delta)x}(A,B)$. Then

$$\text{sign}[B]^{A \to B} = \text{sign}[A] \otimes \delta$$

**Proof.** We outline the proof for $\delta = +$; similar arguments apply to the other signs. For the (possibly posterior) probability distribution over $B$, we have that

$$[\Pr(b)]_{\Pr(X)} =$$
$$[(\Pr(b \mid a) - \Pr(b \mid \bar{a})) \cdot \Pr(a) + \Pr(b \mid \bar{a})]_{\Pr(X)}$$

Now, suppose that sign$[A] = +$. From $S^{?(+)x}(A, B)$ we have that $[\Pr(b \mid a) - \Pr(b \mid \bar{a})]_{\Pr(X)} \geq 0$. From sign$[A] = +$, moreover, we have that $\Pr(a)$ increases. In the current state of the network, therefore, the change in $\Pr(a)$ has a positive effect on $\Pr(b)$ along the arc $A \to B$. We conclude that sign$[B]^{A \to B} = + \otimes +$. Similar observations hold for the situations where sign$[A]$ equals '$-$', '0' or '?'. □

From Proposition 1 and Proposition 2 we have that signs can be propagated over situational influences as if these were regular qualitative influences. To show that the property of composition holds for situational influences, it now suffices to show that for adding a new effect on a node to the prior effects on that node, it is valid to use the $\oplus$-operator from Figure 2. From the property stated in Proposition 2, however, it is readily seen that after propagating a sign over a situational influence, a regular qualitative sign results. A node therefore receives regular signs only, for the composition of which it is valid to use the $\oplus$-operator.

### 3.1.3 Dynamics

We recall that situational signs depend on the state of the network under consideration. The validity of such a sign, therefore, has to be evaluated as observations are entered into the network and, if necessary, the sign has to be updated. In this section, we give a method for maintaining the validity of situational signs.

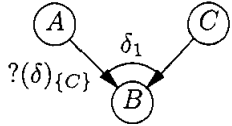

Figure 6: A situational network with $S^{?(\delta)\{C\}}(A, B)$ and $Y^{\delta_1}(\{A, C\}, B)$.

We begin by examining the dynamics of the situational sign of a situational influence $S^{?(\delta)\{C\}}(A, B)$ of a node $A$ on a node $B$, where node $B$ has just two parents $A$ and $C$, as shown in Figure 6. By definition we have that $\delta$ is the sign of the difference $[\Pr(b \mid a) - \Pr(b \mid \bar{a})]_{\Pr(C)}$. This difference can also be expressed as

$$[\Pr(b \mid a) - \Pr(b \mid \bar{a})]_{\Pr(C)} = \Pr(c) \cdot$$
$$(\Pr(b \mid ac) - \Pr(b \mid a\bar{c}) - \Pr(b \mid \bar{a}c) + \Pr(b \mid \bar{a}\bar{c})) +$$
$$\Pr(b \mid a\bar{c}) - \Pr(b \mid \bar{a}\bar{c}) \quad (1)$$

We observe that the above equation expresses the difference $[\Pr(b \mid a) - \Pr(b \mid \bar{a})]_{\Pr(C)}$ as a linear function in $\Pr(c)$. We further observe that the gradient of this function captures the additive synergy of $A$ and $C$ on $B$. The sign of this additive synergy can therefore be exploited for verifying whether or not the sign of the situational influence of $A$ on $B$ retains its validity when observations cause a change in the probability distribution over $C$. If the situational sign does not persist, moreover, the sign of the additive synergy can be used to update the situational sign to a sign that is valid in the new state of the network.

**Lemma 1** *Let $G$ be the digraph of a situational network, with $A, B, C \in V(G)$ and $\pi(B) = \{A, C\}$. Furthermore, let $S^{?(\delta)\{C\}}(A, B)$ and $Y^{\delta_1}(\{A, C\}, B)$. Then,*

$$\delta \leftarrow \delta \oplus (\text{sign}[C] \otimes \delta_1)$$

**Proof.** We outline the proof for sign$[C] = +$; similar arguments apply to the other possible node signs for $C$. Now, suppose that $Y^+(\{A, C\}, B)$. From $Y^+(\{A, C\}, B)$, we have that the gradient in Equation 1 is positive. From sign$[C] = +$, we further have that the probability of $c$ increases. A sign $\delta = +$ for the situational influence will now remain valid. If, on the other hand, we have that $Y^-(\{A, C\}, B)$, then a negative sign for the situational influence will retain its validity. Otherwise, $\delta$ becomes unknown. We conclude that $\delta \leftarrow \delta \oplus (+ \otimes \delta_1)$. □

Without giving a formal proof, we extend our method for updating the sign of a situational influence, as stated in Lemma 1, to a node $B$ having more than two parents. Informally spoken, the sign of the situational influence of node $A$ on node $B$ will remain valid after a change in probability of one or more of the other parents of $B$ if these changes separately serve to preserve the sign.

**Proposition 3** *Let $G$ be the digraph of a situational network, with $A, B, C_1, \ldots, C_n \in V(G), n \geq 1$, and $\pi(B) = \{A, C_1, \ldots, C_n\}$. Furthermore, let $S^{?(\delta)\{C_1,\ldots,C_n\}}(A, B)$ and let $Y^{\delta_i}(\{A, C_i\}, B), i = 1, \ldots, n$. Then,*

$$\delta \leftarrow \delta \oplus_{i=1}^{n} (\text{sign}[C_i] \otimes \delta_i)$$

We illustrate the updating of situational signs by means of our running example.

**Example 4** Consider once again the example Bayesian network from Figure 1 and its extended qualitative abstraction shown in Figure 5. Suppose that the network is just a fragment of a larger network and that observations in other parts can change the prior probability distribution over node $F$. Now, if $\Pr(f)$ decreases, the difference $\Pr(w \mid t) - \Pr(w \mid \bar{t})$ decreases and remains negative. The situational sign '?$(-)_{\{F\}}$' thus retains its validity. If, on the other hand, $\Pr(f)$ increases, the difference will increase. It will change from negative to positive at $\Pr(f) \approx 0.67$. In the qualitative network, where no numerical information is available, the current sign of the situational influence of $T$ on $W$ is no longer known to be valid and is therefore changed to '?(?)'. □

## 3.2 PROVOKERS

The updating of a situational sign, as described in the previous section, can result in a '?(?)' which effectively destroys the possibility of propagating unambiguous information. In this section, we argue that when the variables that underlie the non-monotonicity of an influence, have been observed, the influence reduces to a fixed monotonic one with a regular unambiguous sign. To indicate these variables, we introduce the concept of *provoker set* into the formalism of qualitative probabilistic networks. In addition, we give a method for computing a fixed sign for the reduced influence once values for all variables from the provoker set have been observed, which may re-establish the possibility of propagating unambiguous information.

We recall that a qualitative influence of a node $A$ on a node $B$ is non-monotonic if the difference $\Pr(b \mid ax) - \Pr(b \mid \bar{a}x)$ yields contradictory signs for different combinations of values $x$ for the set $X = \pi(B) \setminus \{A\}$. The provoker set $P \subseteq X$ for the non-monotonic influence now is the set of variables that upon observation serve to reduce the non-monotonic influence to a monotonic one.

**Definition 2** *Let $G$ be the digraph of a qualitative probabilistic network, with $A, B \in V(G), A \to B \in A(G)$, and $P \cup Y = \pi(B) \setminus \{A\}, P \cap Y = \emptyset$. Let $S^?(A, B)$. $P$ is the* provoker set *for the non-monotonic influence of $A$ on $B$ iff for each $P_i \in P$ there exists a combination of values $z$ for the set $Z = P \setminus \{P_i\}$ such that for some $p_{ij}, p_{ik} \in \{p_i, \bar{p}_i\}, p_{ij} \neq p_{ik}$, we have that*

- $\Pr(b \mid ap_{ij}zy) - \Pr(b \mid \bar{a}p_{ij}zy) \geq 0$ *and,*
- $\Pr(b \mid ap_{ik}zy) - \Pr(b \mid \bar{a}p_{ik}zy) \leq 0,$

*for any combination of values $y$ for $Y$.*

We now have, by definition, that for any combination of values $p$ for the set $P$ either $\Pr(b \mid apy) - \Pr(b \mid \bar{a}py) \geq 0$ holds for all combinations of values $y$ for $Y$, or $\Pr(b \mid apy) - \Pr(b \mid \bar{a}py) \leq 0$ holds for all such $y$. Once values for all provokers have been observed, therefore, the non-monotonic influence of $A$ on $B$ reduces to a fixed monotonic one.

To provide for establishing the sign of a reduced non-monotonic influence, we first give a method to compute the sign of a reduced influence of a node $A$ on a node $B$, where $B$ has just one other parent $C$. Note that $\{C\}$ then is the provoker set.

**Lemma 2** *Let $G$ be the digraph of a situational network, with $A, B, C \in V(G)$ and $\pi(B) = \{A, C\}$. Furthermore, let $S^{?(\delta)\{C\}}(A, B)$ and let $Y^{\delta_1}(\{A, C\}, B)$. If a value for $C$ is observed, resulting in $\text{sign}[C]$, then*

$$?(\delta)_{\{C\}} \leftarrow \text{sign}[C] \otimes \delta_1$$

**Proof.** We outline the proof for $\delta_1 = +$; similar arguments apply to the other possible signs for $\delta_1$. From $Y^+(\{A, C\}, B)$, we have that $\Pr(b \mid ac) - \Pr(b \mid \bar{a}c) \geq \Pr(b \mid a\bar{c}) - \Pr(b \mid \bar{a}\bar{c})$. Because the influence of $A$ on $B$ is non-monotonic, this implies that $\Pr(b \mid ac) - \Pr(b \mid \bar{a}c) > 0$ and $\Pr(b \mid a\bar{c}) - \Pr(b \mid \bar{a}\bar{c}) < 0$. We thus have that the influence of $A$ on $B$ is positive after $C = true$ has been observed and negative for $C = false$. We conclude that $?(\delta)_{\{C\}} \leftarrow \text{sign}[C] \otimes \delta_1$. □

Without giving a formal proof, we extend our method as stated in Lemma 2 to nodes with more than two parents.

**Proposition 4** *Let $G$ be the digraph of a situational network, with $A, B, C_1, \ldots, C_n \in V(G), n \geq 1$, and $\pi(B) = \{A, C_1, \ldots, C_n\}$. Let $S^{?(\delta)\{C_1,\ldots,C_n\}}(A, B)$, and let $P = \{P_1, \ldots, P_m\} \subseteq \{C_1, \ldots, C_n\}, 1 \leq m \leq n$, be the provoker set for the influence of $A$ on $B$. Furthermore, let $Y^{\delta_j}(\{A, P_j\}, B), j = 1, \ldots, m$. If each $P_j$ is observed, resulting in $\text{sign}[P_j]$, then*

$$?(\delta)_{\{C_1,\ldots,C_n\}} \leftarrow \oplus_{j=1}^{m} (\text{sign}[P_j] \otimes \delta_j)$$

We observe from the previous proposition that the method for establishing the sign of a reduced influence closely resembles our method for updating a situational sign. The main difference is that for establishing the fixed sign of a reduced influence, its situational sign is of no importance; only the fact that the influence of $A$ on $B$ is non-monotonic is relevant. Any situational sign, and therefore also the sign '?(?)', can thus reduce to an unambiguous regular sign.

So far, we have assumed that each non-monotonic influence has associated a provoker set. Even if this set is not explicitly known, however, Proposition 4 can be applied, by taking $\pi(B) \setminus \{A\}$ for the provoker set. Because the actual set of provokers is a subset of $\pi(B) \setminus \{A\}$, the resulting sign will possibly be weaker, but in any case valid.

## 4 INFERENCE

For inference with a regular qualitative probabilistic network, an efficient algorithm is available, as reviewed in Section 2. Building on the concepts and methods developed in Section 3 we now adapt this algorithm to render it applicable to qualitative networks that include situational influences and extend it to exploit the observation of provokers.

In essence, two modifications to the original algorithm are made. First, upon propagating a sign over a non-monotonic influence, the associated situational sign is used, as indicated by Proposition 2. Also because the situational sign of a non-monotonic influence of a node $A$ on a node $B$ can change whenever an observation causes the state of the network to change, its validity is verified as soon as the probability of another parent of $B$ changes. If necessary, the sign is updated, as indicated by Proposition 3. With just these adaptations, however, it may happen that upon inference a sign is propagated along a situational influence between $A$ and $B$, while the fact that the probability distribution over another parent of $B$ has changed does not become apparent until later during the inference. It may then turn out that the situational sign should have been adapted and that an incorrect sign for the influence of $A$ on $B$ has been used for the propagation. The inference therefore is restarted each time a situational sign is updated. Since a situational sign can change at most once, the number of restarts is limited to the number of situational signs in the network. The second modification to the original algorithm is the exploitation of observations for the provokers of non-monotonic influences. As soon as observations have been entered for all provokers of a non-monotonic influence, the situational sign of the influence is reduced to a regular sign as indicated by Proposition 4.

The adapted algorithm is summarised in pseudo-code in Figure 7. The algorithm uses two pre-defined sets: the set $A_{nm}$ containing all arcs with a situational influence, and the set $C_{nm}$ containing all nodes that are co-parents of a parent exerting a situational influence. The function $A_{nm}(V)$ takes for its argument a node and yields all arcs with a situational influence exerted by the co-parents of this node.

**Example 5** Consider the situational network from Figure 8; the situational sign associated with the influence of node $A$ on node $B$ pertains to the prior state of the network in which no evidence has been entered. For this network, the sets $A_{nm} = \{A \rightarrow B\}$ and $C_{nm} = \{C\}$ are defined. Now suppose that the observation $D = \mathit{false}$ is entered into the network. Prior to propagating this observation, the adapted algorithm checks if, given this new evidence, it can reduce a non-monotonic influence. Since $D \notin C_{nm}$, this is not the case and the propagation proceeds. Node $D$ receives the message '$-$' and updates its node sign to $0 \oplus - = -$. For both its neighbours $A$ and $C$, it then computes the message $- \otimes - = +$. Upon receiving this message, node $C$ updates its node sign to $0 \oplus + = +$. The algorithm now detects that the node sign of an element of $C_{nm}$ has changed and determines that the validity of the situational sign of the influence

$A_{nm} = \{V_i \rightarrow V_j \mid S^{?(\delta)\times}(V_i, V_j)\}$
$C_{nm} = \{V_k \mid V_k \in \pi(V_j) \setminus \{V_i\}, V_i \rightarrow V_j \in A_{nm}\}$
$A_{nm}(V) = \{V_i \rightarrow V_j \mid V_i \rightarrow V_j \in A_{nm}, V_j \in \sigma(V), V_i \neq V\};$

**procedure** Init($Q, OldObs, Obs, sign$):

    $Q \leftarrow$ Fix-Sign($Q$);
    Process-Obs($Q, Obs, sign$).

**function** Fix-Sign($Q$):$Q$

    **if** $Obs \in C_{nm}$
    **then for all** $V_i \rightarrow V_j \in A_{nm}(Obs)$
        **do if** Provokers($S^{?(\delta)\times}(V_i, V_j)) \subseteq OldObs \cup \{Obs\}$
            **then** fix sign for $S(V_i, V_j)$;
        $Q \leftarrow Q$ with fixed signs;
        update $A_{nm}$ and $C_{nm}$;

**procedure** Process-Obs($Q, Obs, sign$):

    **for all** $V_i \in V(G)$ in $Q$
    **do** sign[$V_i$] $\leftarrow$ '0';
    Propagate-Sign($Q, \varnothing, Obs, sign$).

**procedure** Propagate-Sign($Q, trail, to, message$):

    sign[$to$] $\leftarrow$ sign[$to$] $\oplus$ $message$;
    $trail \leftarrow trail \cup \{to\}$;
    Determine-Effect-On($Q, to$);
    **for** each neighbour $V_i$ of $to$ in $Q$
    **do** $linksign \leftarrow$ sign of influence between $to$ and $V_i$;
        $message \leftarrow$ sign[$to$] $\otimes$ $linksign$;
        **if** $V_i \notin trail$ and sign[$V_i$] $\neq$ sign[$V_i$] $\oplus$ $message$
        **then** Propagate-Sign($Q, trail, V_i, message$).

**procedure** Determine-Effect-On($Q, V_i$):

    **if** $V_i \in C_{nm}$
    **then for all** $V_j \rightarrow V_k \in A_{nm}(V_i)$
        **do** Verify-Update($S^{?(\delta)\times}(V_j, V_k)$);
        **if** a $\delta$ changes
        **then** $Q \leftarrow Q$ with adapted signs;
            **return** Process-Obs($Q, Obs, sign$).

Figure 7: The adapted sign-propagation algorithm.

of $A$ on $B$ needs to be verified. Since $+ \oplus (+ \otimes +) = +$ equals the current regular part of the situational sign, no updating is required and the inference continues. Node $C$ sends the message $+ \otimes + = +$ to node $B$, causing $B$ to update its node sign to $0 \oplus + = +$. $B$ does not send the message it has received from $C$ to node $A$ because $C$ and $A$ are independent on the trail $C, B, A$. Upon receiving the message '+' from node $D$, node $A$ now updates its node sign to $0 \oplus + = +$ and subsequently sends the message $+ \otimes + = +$ to $B$, thereby using the situational sign of the influence of $A$ on $B$. Since node $B$ does not need to change its node sign after receiving this message, the inference halts. Entering the observation $D = \mathit{false}$ into the network therefore results in the node signs '+', '+', '+' and '−' for $A$, $B$, $C$ and $D$, respectively.

Now suppose that the observation $D = \mathit{true}$ is entered instead. Node $D$ will then compute the message

$+ \otimes - = -$ for both $A$ and $C$. Upon receiving this message, node $C$ updates its node sign to $0 \oplus - = -$. Verification of the validity of the situational sign of the influence of node $A$ on node $B$ now reveals that $- \oplus (+ \otimes +) = ?$ differs from the current regular part of the sign. The situational sign is therefore updated to '$?(?)_{\{C\}}$' and inference starts anew with the adapted network. The inference now results in the node signs '$-$', '$?$', '$-$' and '$+$' for $A$, $B$, $C$ and $D$, respectively.

After the observation $D = true$ has been entered and propagated, the network specifies the situational sign '$?(?)_{\{C\}}$' for the influence of node $A$ on node $B$. Now suppose that the observation $C = false$ is subsequently entered into the network. The adapted algorithm again first checks if it can reduce a non-monotonic influence to a monotonic one. It now finds that the provoker for the non-monotonic influence of $A$ on $B$ has been observed. Using the sign of the additive synergy of $A$ and $C$ on $B$, it computes the sign $- \otimes + = -$ for the reduced influence and replaces the situational sign '$?(?)_{\{C\}}$' by the regular sign '$-$'. The inference now proceeds as in a regular qualitative network. □

## 5 CONCLUSIONS

Qualitative probabilistic networks model non-monotonic relationships between their variables by means of the ambiguous sign '?'. The presence of influences with ambiguous signs typically leads to ambiguous, and thus uninformative, results upon inference. In this paper we demonstrated that more informative inference results can be obtained in the presence of non-monotonicity. We extended the formalism of qualitative networks with situational signs that capture qualitative information about the current effect of non-monotonic influences. We justified the use of these signs upon inference and provided for maintaining their validity. Furthermore, we characterised the provokers of a non-monotonic influence. We showed that observation of these provokers reduces the influence to a monotonic one and we gave a method for establishing a sign for the reduced influence. To conclude, we adapted the standard sign-propagation algorithm to render it applicable to qualitative networks with situational signs and extended it to exploit the observation of provokers. We showed that the new algorithm may effectively forestall ambiguous results upon inference. To summarise, we strengthened the expressiveness of a qualitative network by adding and exploiting information about non-monotonic influences. Recently, other research also focused on enhancing the formalism of qualitative networks, for example by introducing a notion of strength [6]. In the future we will investigate how these different enhancements can be integrated to arrive at an even more powerful framework for qualitative probabilistic reasoning.

### Aknowledgment

This research was supported by the Netherlands Organisation for Scientific Research (NWO).

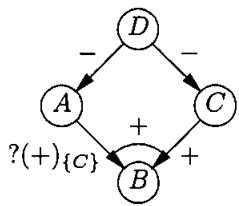

Figure 8: A network with $S^-(D,A)$, $S^-(D,C)$, $S^{?(+)\{C\}}(A,B)$, $S^+(C,B)$ and $Y^+(\{A,C\},B)$.